\documentclass[11pt,a4paper]{article}
\usepackage[hyperref]{naaclhlt2019}
\usepackage{times}
\usepackage{latexsym}

\usepackage{url}

\usepackage{graphicx}
\usepackage{color}
\usepackage{multirow}
\usepackage{amssymb}
\usepackage{array}
\usepackage[normalem]{ulem}

\usepackage{float}

\usepackage[compact]{titlesec}
\titlespacing{\section}{0pt}{1ex}{1ex}
\titlespacing{\subsection}{0pt}{1ex}{0ex}
\titlespacing{\subsubsection}{0pt}{0.5ex}{0ex}

\aclfinalcopy 
\def\aclpaperid{1223} 

\newcommand{\citea}[1]{\citeauthor{#1} (\citeyear{#1})}

\title{Topic Spotting using Hierarchical Networks with Self Attention}

\author{ Pooja Chitkara$^{1}$, 
    Ashutosh Modi$^{1}$,
  Pravalika Avvaru$^{1}$, 
  Sepehr Janghorbani$^{1,2}$, 
  Mubbasir Kapadia$^{1,2}$ \\ 
  {$^1$Disney Research, } 
  {$^{2}$Rutgers University}\\
  {\tt pchitkar@andrew.cmu.edu}\\
  {\tt ashutosh.modi@disneyresearch.com} \\
  {\tt pavvaru@andrew.cmu.edu} \\
  {\tt sepehr.janghorbani@rutgers.edu} \\
  {\tt mubbasir.kapadia@rutgers.edu}
  \\}

\date{}

\begin{document}
\maketitle
\begin{abstract}
Success of deep learning techniques have renewed the interest in development of dialogue systems. However, current systems struggle to have consistent long term conversations with the users and fail to build rapport. Topic spotting, the task of automatically inferring the topic of a conversation, has been shown to be helpful in making a dialog system more engaging and efficient. We propose a hierarchical model with self attention for topic spotting. Experiments on the Switchboard corpus show the superior performance of our model over previously proposed techniques for topic spotting and deep models for text classification. Additionally, in contrast to offline processing of dialog, we also analyze the performance of our model in a more realistic setting i.e. in an online setting where the topic is identified in real time as the dialog progresses. Results show that our model is able to generalize even with limited information in the online setting.
\end{abstract}


\section{Introduction} \label{sec:introduction}

Recently,  a number of commercial conversation systems have been introduced e.g. Alexa, Google Assistant, Siri, Cortana, etc. Most of the available systems perform well on goal-oriented conversations which spans over few utterances in a dialogue. However, with longer conversations (in open domains), existing systems struggle to remain consistent and tend to deviate from the current topic during the conversation. This hinders the establishment of long term social relationship with the users \cite{dehn2000impact}. In order to have coherent and engaging conversations with humans, besides other relevant natural language understanding (NLU) techniques \cite{jokinen2009spoken}, a system, while responding, should take into account the topic of the current conversation i.e. Topic Spotting. 

Topic spotting has been shown to be important in commercial dialog systems \cite{bost2013multiple,jokinen2002adaptive} directly dealing with the customers. Topical information is useful for speech recognition systems \cite{iyer1999modeling} as well as in audio document retrieval systems \cite{hazen2007topic,hazen2011topic}. Importance of topic spotting can be gauged from the work of Alexa team \cite{corr2018}, who have proposed topic based metrics for evaluating the quality of conversational bots. The authors empirically show that topic based metrics correlate with human judgments. 

Given the importance of topical information in a dialog system, this paper proposes self attention based hierarchical model for predicting topics in a dialog. We evaluate our model on Switchboard (SWBD) corpus \cite{godfrey1992switchboard} and show that our model supersedes previously applied techniques for topic spotting. We address the evaluative limitations of the current SWBD corpus by creating a new version of the corpus referred as SWBD2. We hope that SWBD2 corpus would provide a new standard for evaluating topic spotting models. We also experiment with an online setting where we examine the performance of our topic classifier as the length of the dialog is varied and show that our model can be used in a real time dialog system as well. 


\section{Related Work} \label{sec:relatedWork}

\begin{figure*}[t]
\centering
  \includegraphics[scale=0.33]{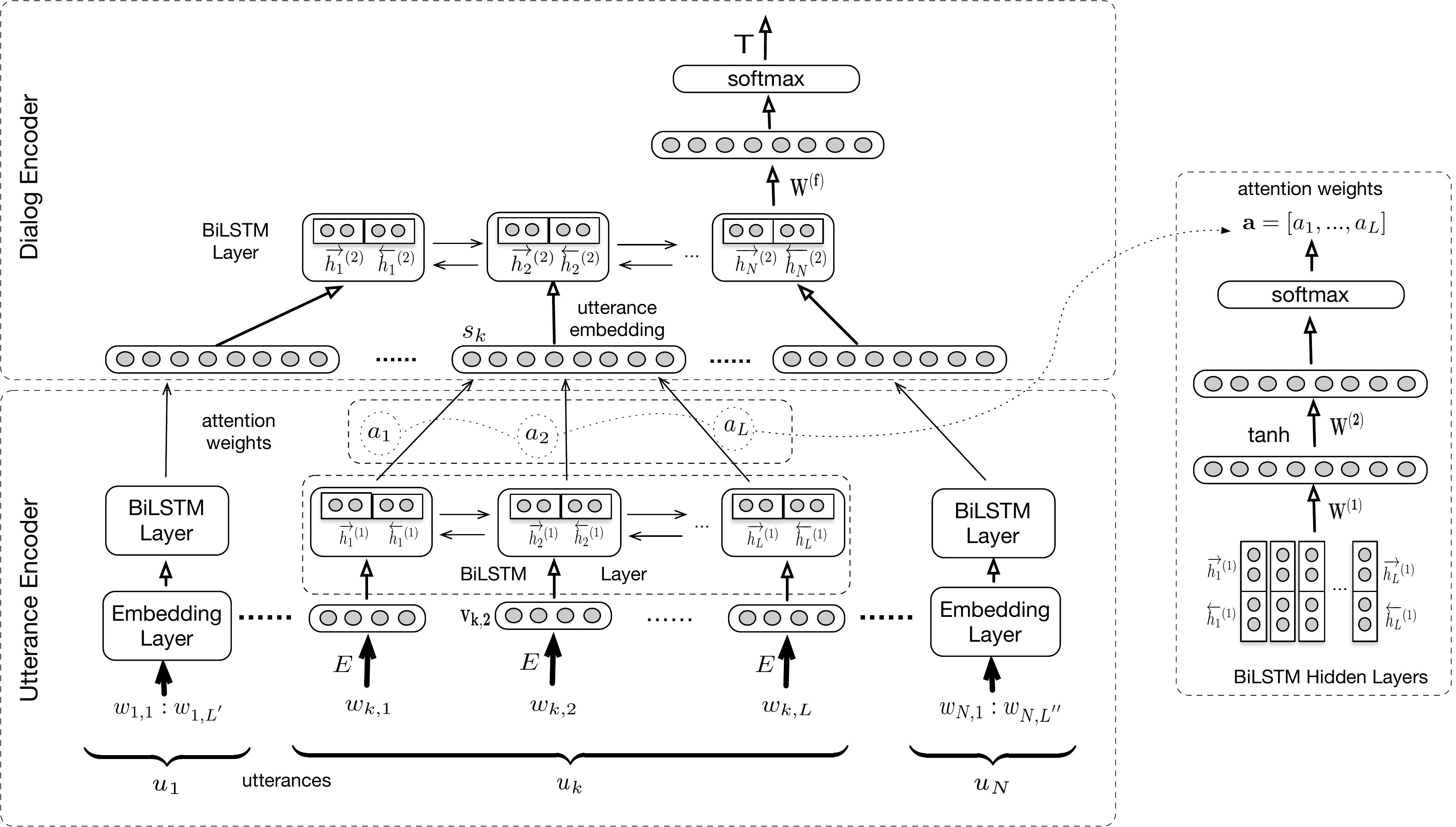}
  \caption{Model Architecture}
  \label{fig:model}
\end{figure*}

Topic spotting is the task of detecting the topic of a dialog \cite{hazen2007topic}. Topic spotting has been an active area of research over the past few decades both in the NLP community as well as in the speech community. In this section we briefly outline some of the main works in this area. For a detailed survey of prior research in this area, the reader is referred to \citea{hazen2011topic}. 

Most of the methods proposed for topic spotting use features extracted from transcribed text as input to a classifier (typically Na\"ive Bayes or SVM ). Extracted features include: Bag of Words (BoW), TF-IDF \cite{sparck1972statistical,schutze2008introduction}, n-grams, and word co-occurrences \cite{hazen2011topic,myers2000boosting}.  Some approaches (in addition to word co-occurrences features) incorporate background world knowledge using Wikipedia \cite{gupta2007topic}. In our work, we do not explicitly extract the features but learn these during training. Moreover, unlike previous approaches, we explicitly model the dependencies between utterances via self attention mechanism and hierarchical structure.

Topic spotting has been explored in depth in the speech processing community (see for example, \citea{wright1996statistical}; \citea{kuhn1997approaches}; \citea{noth1997frame}; \citea{theunissen2002phonene}). Researchers in this community have attempted to predict the topic directly from the audio signals using phoneme based features. However, the performance of word based models supersedes those of audio models \cite{hazen2007topic}. 

Recently, there has been lot of work in deep learning community for text classification \cite{KalchbrennerGB14,zhang2015character,Lai2015,Lin2015HierarchicalRN,tang2015document}. 
These deep learning models use either RNN-LSTM  based neural networks \cite{hochreiter1997long} or CNN based neural networks \cite{kim2014convolutional} for learning representation of words/sentences. We follow similar approach for topic spotting. 
Our model is related to the Hierarchical Attention Network (HN-ATT) model proposed by \citea{yang2016hierarchical} for document classification. HN-ATT models the document hierarchically by composing words (with weights determined by first level of attention mechanism) to get sentence representations and then combines the sentence representations with help of second level attention to get document representation which is then used for classification. 

The aim of this paper is not to improve text classification but to improve topic spotting. Topic spotting and text classification differ in various aspects. We are among the first to show the use of hierarchical self attention (HN-SA) model for topic spotting. It is natural to consider applying text classification techniques for topic spotting. However, as we empirically show in this paper, text classification techniques do not perform well in this setting. Moreover, for the dialog corpus simple BoW approaches perform better than more recently proposed HN-ATT model \cite{yang2016hierarchical}.

\section{Hierarchical Model with Self Attention}  \label{sec:model}

We propose a hierarchical model with self attention (HN-SA) for topic spotting. 
We are given a topic label for each dialog and we want to learn a model mapping from space of dialogues to the space of topic labels. We learn a prediction model by minimizing Negative Log Likelihood ($\mathcal{NLL}$) of the data. 

\subsection{Model Architecture}

We propose a hierarchical architecture as shown in Figure \ref{fig:model}. An  \textit{utterance encoder} takes each utterance in the dialog and outputs the corresponding utterance representation. A \textit{dialog encoder} processes the utterance representations to give a compact vector representation for the dialog which is used to predict the topic of the dialog. 

\noindent\textbf{Utterance Encoder:} Each utterance in the dialog is processed sequentially using single layer Bi-directional Long Short Term Memory (BiLSTM) \cite{dyer2015transition} network and self-attention mechanism \cite{vaswani2017attention} to get the  utterance representation. In particular, given an utterance with one-hot encoding for the tokens, $u_{k} = \{\mathbf{w_{k,1}, w_{k,2},....,w_{k,L}}\}$, each token is mapped to a vector  $\mathbf{v_{k,i}} = \mathbf{E} \mathbf{w_{k,i}} \ \ ;i=1,2,...L$ using pre-trained embeddings (matrix $\mathbf{E}$). 


Utterance representation ($\mathbf{s_{k}} = \mathbf{a}^{T} \mathbf{H^{(1)}}$) is the weighted sum of the forward and backward direction concatenated hidden states at each step 
of the BiLSTM ($\mathbf{H^{(1)}} = [\mathbf{h_{1}^{(1)}},....,\mathbf{h_{L}^{(1)}}]^{T}$ where $\mathbf{h_{i}^{(1)}} = [\overrightarrow{\mathbf{{h_{i}}}}^{(1)}:\overleftarrow{\mathbf{h_{i}}}^{(1)}] = \mathbf{BiLSTM}(\mathbf{v_{k,i}})$ ). The weights of the combination ($\mathbf{a}  = \textrm{softmax}(\mathbf{h^{(2)}_{a}})$) are determined using self-attention mechanism proposed by \citea{vaswani2017attention} by measuring the similarity between the concatenated hidden states ($\mathbf{h^{(2)}_{a}} = \mathbf{W_{a}^{(2)}} \mathbf{h^{(1)}_{a}} + \mathbf{b_{a}^{(2)}}$ and $\mathbf{h^{(1)}_{a}} = \textrm{tanh} ( \mathbf{W_{a}^{(1)}} \mathbf{H^{(1)}} + \mathbf{b_{a}^{(1)}})$) at each step in the utterance sequence.  Self-attention computes the similarity of a token in the context of an utterance and thus, boosts the contribution of some keywords to the classifier. It also mitigates the need for a second layer of attention at a dialog level reducing the number of parameters, reducing the confusion of the classifier by not trying to reweigh individual utterances and reducing the dependence on having all utterances (full future context) for an accurate prediction. A simple LSTM based model (HN) and HN-ATT perform worse than the model using self attention (\textsection\ref{sec:results}), indicating the crucial role played by self-attention mechanism.




\noindent\textbf{Dialog Encoder:} Utterance embeddings (representations) are sequentially encoded by a second single layer BiLSTM to get the dialog representation ($\mathbf{h_{k}^{(2)}} = [\overrightarrow{\mathbf{{h_{k}}}}^{(2)}:\overleftarrow{\mathbf{h_{k}}}^{(2)}] = \mathbf{BiLSTM}(\mathbf{s_{k}}) \ \  ;k=1,2,...N$). Bidirectional concatenated hidden state corresponding to the last utterance (i.e. last step of BiLSTM) is used for making a prediction via a linear layer followed by softmax activation ($p(\mathsf{T} | \mathsf{D}) = \textrm{softmax}(\mathbf{h_{D}})$ where $\mathbf{h_{D}} = \mathbf{W_{f}} \mathbf{h_{N}^{(2)}}$).



\begin{table}[t]
\footnotesize
\resizebox{\columnwidth}{!}{%
\begin{tabular}{ccccccc}
\hline
      & \multicolumn{2}{c}{\# Dialogues} & \multicolumn{2}{c}{\# Topics} & \multicolumn{2}{c}{Avg. \# Utterances} \\ \hline 
      & \textbf{SWBD}  & \textbf{SWBD2}  & \textbf{SWBD} & \textbf{SWBD2} & \textbf{SWBD}  & \textbf{SWBD2}         \\ \hline \hline
Train & 1024           & 877             & 66            & 42            & 192.27          & 194.33         \\
Dev   & 112            & 49              & 48            & 33            & 180.52          & 177.02         \\
Test  & 19             & 98              & 12            & 42            & 237.58          & 201.97        
\end{tabular}
}
\caption{Corpus statistics for both versions of SWBD}
\label{table:swbdStats}
\end{table}

\section{Experimental Setup}  \label{sec:experiments}

As in previous work (\textsection{\ref{sec:relatedWork}}), we use Switchboard (SWBD) \cite{godfrey1992switchboard} corpus for training our model. SWBD is a corpus of human-human conversations, created by recording (and later transcribing) telephonic conversations between two participants who were primed with a topic. Table \ref{table:swbdStats} gives the corpus statistics. Topics in SWBD range over a variety of domains, for example, politics, health, sports, entertainment, hobbies, etc., making the task of topic spotting challenging. 

Dialogues in the test set of the original SWBD cover a limited number of topics (12 vs 66). The test set is not ideal for evaluating topic spotting system. We address this shortcoming by creating a new split and we refer to this version of the corpus as \textit{SWBD2}. The new split provides opportunity for more rigorous evaluation of a topic spotting system. SWBD2 was created by removing infrequent topics (\textless\ 10 dialogues) from the corpus and then randomly moving dialogues between the train/development set and the test set, in order to have instances of each topic in the test set. 
The majority class baseline in SWBD2 is around 5\%.

In transcribed SWBD corpus some punctuation symbols such as \#, ?, have special meanings and non-verbal sounds have been mapped to special symbols e.g. \textless Laughter\textgreater. To preserve the meanings of special symbols we performed minimal preprocessing. Dialog Corpora is different from text classification corpora (e.g. product reviews). If we roughly equate a dialog to a document and an utterance to a sentence, dialogs are very long documents with short sentences. Moreover, the vocabulary distribution in a dialog corpus is fundamentally different, e.g. presence of back-channel words like `uhm' and `ah'.

\noindent\textbf{Model Hyper-parameters:} We use GloVe embeddings \cite{pennington2014glove} with dimensionality of 300. The embeddings are updated during training. Each of the LSTM cell in the utterance and dialog encoder uses hidden state of dimension 256. The weight matrices in the attention network have dimension of 128. The hyper-parameters were found by experimenting with the development set. We trained the model by minimizing the cross-entropy loss using Adam optimizer \cite{kingma2014adam} with an initial learning rate of 0.001. The learning rate was reduced by half when development set accuracy did not change over successive epochs. Model took around 30 epochs to train.   



\section{Experiments and Results}  \label{sec:results}

\begin{table}[]
\footnotesize
\begin{tabular}{lrr}
\hline \hline
\textbf{Models}                                                        & \textbf{SWBD}  & \textbf{SWBD2} \\ \hline\hline
BoW + Logsitic                                                     & 78.95 & 87.76    \\ 
BoW + SVM                                                          & 73.68 & \textbf{90.82}    \\ 
Bigram + SVM                                                                  & 52.63 & 79.59    \\ 
BoW + TF-IDF + Logistic                                                      & 52.63 & 81.63    \\ 
nGram + Logistic                                                            & 52.63 & 78.57    \\ 
nGram + TF-IDF + Logistic                                                    & 57.89 & 87.76    \\ 
Bag of Means + Logistic                                                     & 78.95 & 87.76    \\ \hline
Avg. Skipgram + Logistic  & 26.32 & 59.18    \\ 
Doc2Vec + SVM                      & 73.68 & 86.73    \\ 
HN                                                       &   31.58    & 54.08         \\ 
HN-ATT \cite{yang2016hierarchical}                             & 73.68 & 85.71    \\ 
CNN \cite{kim2014convolutional} & 84.21 & 93.87\\
\hline \hline
HN-SA (our model)                                                                   & \textbf{89.47} & \textbf{95.92}    \\ \hline \hline
\end{tabular}
\caption{Accuracy (in \%) of our model and other text classification models on both versions of SWBD.}
\label{table:result}
\vspace{-6mm}
\end{table}

We compare the performance of our model (Table  \ref{table:result}) with traditional Bag of Words (BoW), TF-IDF, and n-grams features based classifiers. We also compare against averaged Skip-Gram \cite{mikolov2013distributed}, Doc2Vec \cite{le2014distributed}, CNN \cite{kim2014convolutional}, Hierarchical Attention (HN-ATT) \cite{yang2016hierarchical} and hierarchical network (HN) models.  HN it is similar to our model HN-SA but without any self attention. 

\begin{figure*}[h!]
\centering
  \includegraphics[scale=0.18]{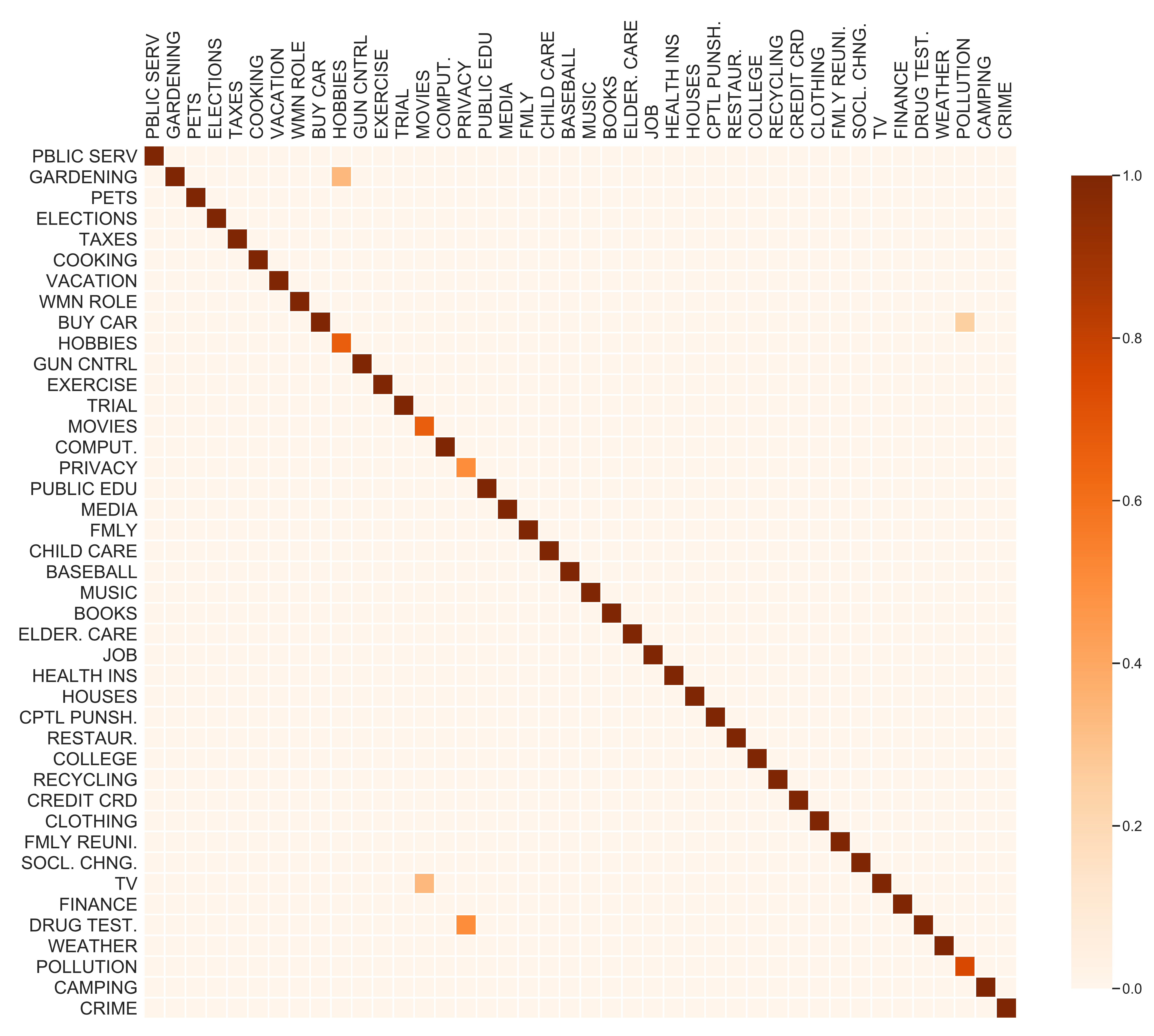}
  \caption{Normalized Confusion Matrix in form of heatmap for model predictions on SWBD2. Vertical axis is the target class.}
  \label{fig:heatmap}

\centering
  \includegraphics[scale=0.37]{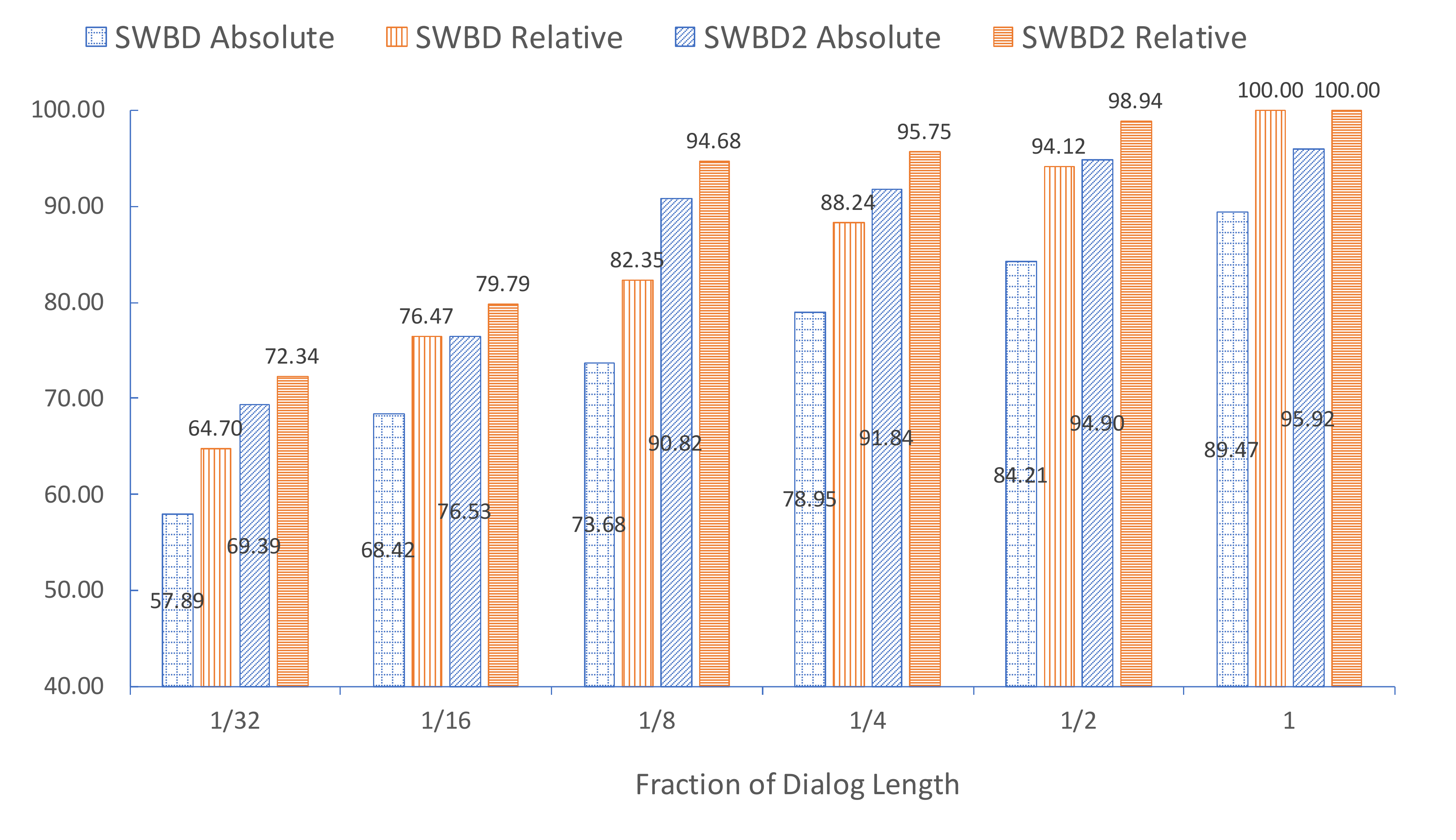}
   \vspace{-5mm}
  \caption{Effect of Dialog Length on Accuracy. Plot shows both the absolute accuracy and relative accuracy (w.r.t. full model) for different fractions of the data.}
  \label{fig:subdialog}
\end{figure*}

\noindent\textbf{Analysis:} As is evident from the experiments on both the versions of SWBD, our model (HN-SA) outperforms traditional feature based  topic spotting models and deep learning based document classification models. It is interesting to see that simple BoW and n-gram baselines are quite competitive and outperform some of the deep learning based document classification model. Similar observation has also been reported by \citea{mesnil2014ensemble} for the task of sentiment analysis. The task of topic spotting is arguably more challenging than document classification. In the topic spotting task, the number of output classes (66/42 classes) is much more than those in document classification (5/6 classes), which is done mainly on the texts from customer reviews.  Dialogues in SWBD have on an average 200 utterances and are much longer texts than customer reviews. Additionally, the number of dialogues available for training the model is significantly lesser than customer reviews. 
We further investigated the performance on SWBD2 by examining the confusion matrix of the model. Figure \ref{fig:heatmap} shows the heatmap of the normalized confusion matrix of the model on SWBD2. For most of the classes the classifier is able to predict accurately. However, the model gets confused between the classes which are semantically close (w.r.t. terms used) to each other, for example, the model gets confused between pragmatically similar topics e.g. HOBBIES vs GARDENING, MOVIES vs TV PROGRAMSâ, RIGHT TO PRIVACY vs DRUG TESTING. 

\noindent\textbf{Online Setting:} In an online conversational system, a topic spotting model is required to predict the topic accurately  and as soon as possible during the dialog. We investigated the relationship between dialog length (in terms of number of utterances) and accuracy. This would give us an idea about how many utterances are required to reach a desirable level of accuracy. For this experiment, we varied the length of the dialogues from the test set that was available to the model for making prediction. We created sub-dialogues of length starting with $1/32$ of the dialog length and increasing it in multiples of 2, up to the full dialog. Figure \ref{fig:subdialog} shows both the absolute accuracy and the accuracy relative to that on the full dialog. With just a few (3.125\%) initial utterances  available, the model is already 72\% confident about the topic. This may be partly due to the fact that in a discussion, the first few utterances explicitly talk about the topic. However, as we have seen, since SWBD covers many different topics which are semantically close to each other but are assigned distinct classes, it is equally challenging to predict the topic with the same model. By the time the system has processed half the dialog in SWBD2 it is already within 99\% accuracy of the full system. The experiment shows the possibility of using the model in an online setting where the model predicts the topic with high confidence as the conversation progresses.   



\section{Conclusion and Future Work}  \label{sec:conclusion}

In this paper we presented a hierarchical model with self attention for topic spotting. The model outperforms the conventional topic spotting techniques as well as deep learning techniques for text classification. We empirically show that the proposed model can also be used in an online setting. We also introduced a new version of SWBD corpus: SWBD2. We hope that it will serve as the new standard for evaluating topic spotting models.
Moving forward, we would like to explore a more realistic multi-modal topic spotting system. Such a system should fuse two modalities: audio and transcribed text to make topic predictions. 



\section*{Acknowledgments}

We would like to thank anonymous reviewers for their insightful comments. Mubbasir Kapadia has been funded in part by NSF IIS-1703883, NSF S\&AS-1723869, and DARPA SocialSim-W911NF-17-C-0098.


\bibliography{naaclhlt2019}
\bibliographystyle{acl_natbib}

\end{document}